\documentclass[10pt,twocolumn,oneside,letterpaper,conference]{IEEEtran}
\usepackage{cite}
\usepackage{graphicx}
\usepackage{amsmath}
\usepackage{times}
\usepackage{latexsym}
\usepackage{graphicx}
\usepackage{bm}
\usepackage{amssymb}
\usepackage[center]{caption2}
\usepackage{stfloats}
\usepackage{cases}
\usepackage{array}
\usepackage{setspace}
\usepackage{fancyhdr}
\usepackage{fancyhdr}
\usepackage{algorithm}
\usepackage{algpseudocode}
\usepackage{textcomp}
\usepackage{wasysym}
\usepackage{balance} 
\usepackage{url}
\usepackage{xcolor}
\usepackage{bmpsize}
\graphicspath{{./figures/}}

\allowdisplaybreaks[4]

\newcaptionstyle{mystyle1}{%
	\centering TABLE  \captiontext \par}
\captionstyle{mystyle1}
\newcaptionstyle{mystyle2}{%
	\captionlabel $.$ \: \captiontext \par}
\captionstyle{mystyle2}
\newcaptionstyle{mystyle3}{%
	\captionlabel $.$ \: \captiontext \par}
\captionstyle{mystyle3}

\begin{document}
	\title{Content Popularity Prediction in Fog-RANs: A Clustered Federated Learning Based Approach}
	\author{
		\IEEEauthorblockN{Zhiheng Wang$^{1}$, Yanxiang Jiang$^{1,2,*}$,
			Fu-Chun Zheng$^{1,2}$, Mehdi Bennis$^3$, and Xiaohu You$^1$}
		\IEEEauthorblockA{$^1$National Mobile Communications Research Laboratory,
			Southeast University, Nanjing 210096, China.\\
			$^2$School of Electronic and Information Engineering, Harbin Institute of Technology, Shenzhen 518055, China.\\
			$^3$Centre for Wireless Communications, University of
			Oulu, Oulu 90014, Finland.\\
			E-Mail: $\{$220200860@seu.edu.cn, yxjiang@seu.edu.cn, fzheng@ieee.org, mehdi.bennis@oulu.fi, xhyu@seu.edu.cn$\}$
	}}
	\date{}
	
	\maketitle

	\begin{abstract}
		In this paper, the content popularity prediction problem in fog radio access networks (F-RANs) is investigated. Based on \textit{clustered federated learning}, we propose a novel mobility-aware popularity prediction policy,
		which integrates content popularities in terms of local users and mobile users. 
		For local users, the content popularity is predicted by learning the hidden representations of local users and contents.
		Initial features of local users and contents are generated by incorporating neighbor information with self information. Then, dual-channel neural network (DCNN) model is introduced to learn the hidden representations by producing deep latent features from initial features.
		For mobile users, the content popularity is predicted via user preference learning.
		In order to distinguish regional variations of content popularity, clustered federated learning (CFL) is employed, which enables fog access points (F-APs) with similar regional types to benefit from one another and provides a more specialized DCNN model for each F-AP.
		Simulation results show that our proposed policy achieves significant performance improvement over the traditional policies.
	\end{abstract}
    \begin{IEEEkeywords}
    	F-RANs, popularity prediction, clustered federated learning, user preference, mobility-aware.
    \end{IEEEkeywords}

    \section{Introduction}   
    With the unprecedented growth of smart devices and mobile application services, tremendous number of problems emerge in wireless networks, especially the congestion caused by the notable data traffic pressure on capacity-limited backhaul links. Fog radio access network (F-RAN) has been introduced as a promising architecture to alleviate the traffic burden on backhaul links via densely deployed fog access points (F-APs) at network edges \cite{Intro1}. F-APs can cache popular contents to satisfy user requests \cite{Hu}.
    As a consequence of constrained caching capacity and computing resources, F-APs need to predict the future content popularity to decide which content to prefetch for the purpose of improving caching efficiency \cite{Intro2}. 
    
    Continuous research efforts have been devoted to predicting content popularity. Except for the traditional methods such as least recently used (LRU) and least frequently used (LFU), machine learning has emerged as the most popular approach to predict content popularity. 
    In \cite{Ma}, an online content popularity prediction was proposed to track the popularity in time by leveraging the content features and user preference.
    In \cite{Feng}, a content classifier was constructed and content popularity was predicted through training a simplified bidirectional long short-term memory (Bi-LSTM) network for every content class.    
    In \cite{Wu}, the authors proposed a popularity prediction policy based on federated learning (FL) by leveraging user preference in adaptively partitioned context spaces.
    In \cite{Tao}, the authors utilized Bayesian learning to get the content request pattern modeled by a Gaussian process based Poisson regressor.
    In \cite{FLCF}, a proactive content caching scheme based on FL was proposed to ensure the privacy of user data.
    In \cite{pLSA1}, user request behavior was modeled by probabilistic latent semantic analysis (pLSA) to predict content popularity.
    However, user mobility is neglected in these existing works except for \cite{Ma} and \cite{Feng}.
    In \cite{Feng} and \cite{Wu}, the edge caching scenario is considered in a specific region, where all F-APs are thought to have the same content popularity.
    Confronted with the incongruent data distributions of F-APs with different regional types, a single popularity prediction model for all F-APs can hardly achieve satisfactory performance.
    Moreover, most works ignore the hidden representations of users and contents which contribute to predicting content popularity.

    Motivated by the aforementioned discussions, we propose a mobility-aware popularity prediction policy based on clustered federated learning (CFL). 
    On one hand, the content popularity in terms of local users is predicted by utilizing a dual-channel nerual network (DCNN) model. 
    The DCNN model of each F-AP is learned automatically with historical request records and the initial features of local users and contents constructed by neighbor selection.
    Moreover, CFL is employed to provide a more specialized model for each F-AP.
    On the other hand, the content popularity in terms of mobile users is predicted via user preference learning.
    Finally, the content popularities in terms of both local users and mobile users are integrated.
    
    The rest of this paper is organized as follows. In Section \uppercase\expandafter{\romannumeral2}, the system model is presented. The proposed popularity prediction policy is described in Section \uppercase\expandafter{\romannumeral3}. Simulation results are shown in Section \uppercase\expandafter{\romannumeral4}. Final conclusions are drawn in Section \uppercase\expandafter{\romannumeral5}.
    
    \section{System Model}
\begin{figure}[!t]
	   \centering
	   \includegraphics[height=4cm]{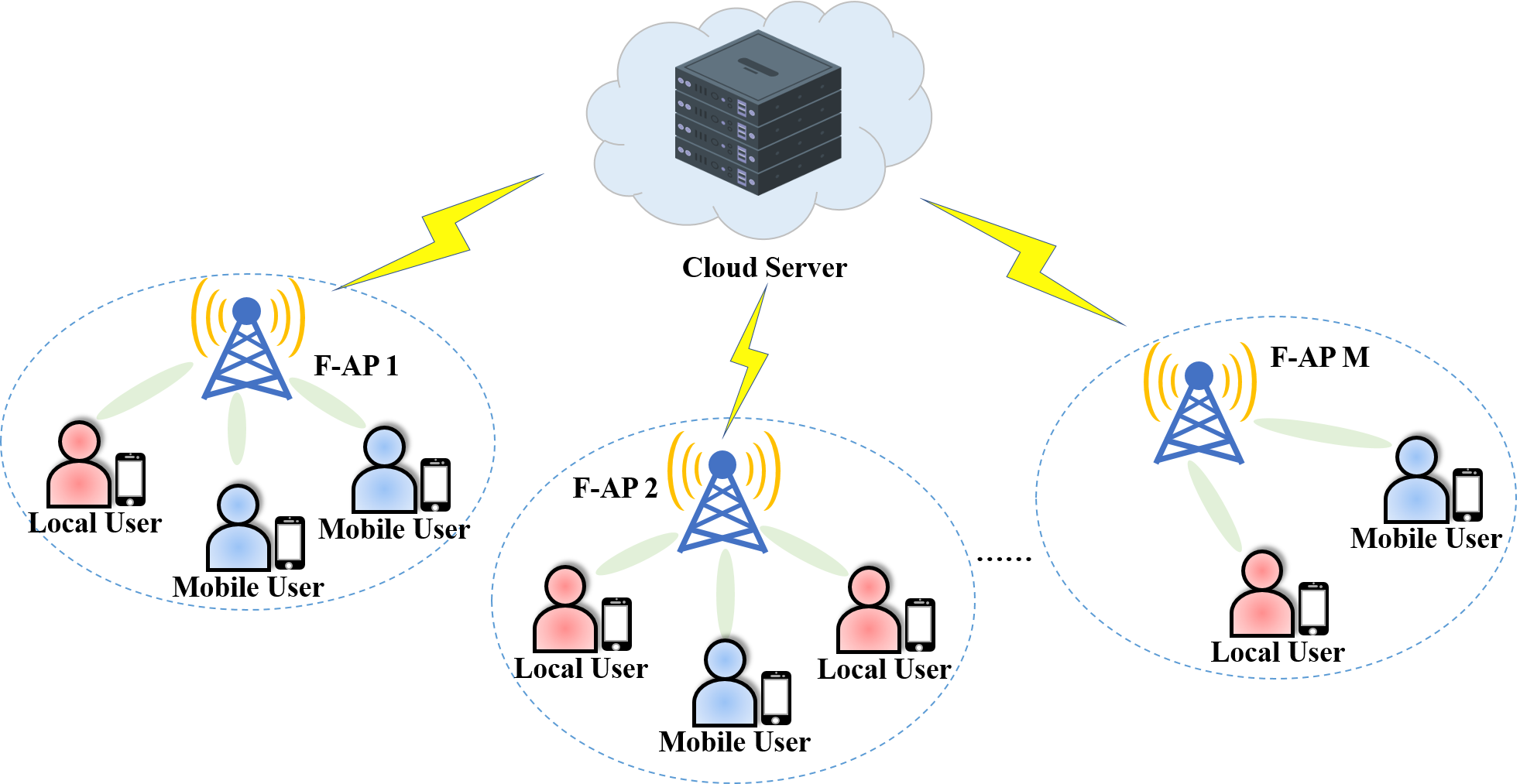} 
	   \caption{Illustration of the edge caching scenario in F-RANs.}  \label{fig:System model}
      \end{figure}

       As illustrated in Fig. \ref{fig:System model}, the edge caching scenario is considered with $M$ F-APs and numerous users. Let $\mathcal{M}=\left\{1, 2, \ldots, m, \ldots, M\right\}$ denote the set of F-APs. It is assumed that each F-AP can serve users within its own coverage area by fetching contents from its storage-limited local cache \cite{Peng}.
       We assume that according to user mobility, users can be categorized into two types: local users and mobile users \cite{Mobile2}. Every local user is associated with a specific F-AP and will not move to any other F-AP within the considered time period in the future, whereas mobile users may randomly visit a certain F-AP at some moment and then keep associated with the F-AP until leaving its coverage area. Let $\mathcal{L}^{m}=\left\{l^{m}_{1}, l^{m}_{2}, \dots, l^{m}_{u}, \dots \right\}$ denote the set of local users associated with F-AP $m$, and $\mathcal{K}=\left\{k_{1}, k_{2}, \dots, k_{u}, \dots \right\}$ the set of mobile users. Each F-AP continuously monitors mobile users in its coverage area.
       Without loss of generality, we assume that all the F-APs have the same storage space which can cache up to $\varphi$ contents from the content library $\mathcal{I}=\left\{1, 2, \dots, i, \dots, I\right\}$ \cite{CuiTrans}. If a user requests content $i$ which has been stored in its associated F-AP, the user can fetch it from the local cache of the F-AP and a cache hit event occurs. Otherwise, the F-AP needs to fetch the content from the cloud server or its neighboring F-APs \cite{PengTrans}.
       
       Content popularity prediction in a given F-AP should consider both local users and mobile users.
       On one hand, content popularity in terms of local users (referred to as local popularity) varies in F-APs with different regional types.
       Therefore, a more specialized model for each F-AP to predict local popularity is required.
       On the other hand, content popularity in terms of mobile users (referred to as mobile popularity) changes continuously due to user mobility \cite{FengTrans}. 
       In addition, F-APs cannot access user information and historical request records of mobile users due to privacy reasons.
       Therefore, user preference learning is required to be implemented independently by each mobile user to predict mobile popularity.
       Since the request probability of users reveals the level of acceptability of contents, we define the content popularity in a given F-AP as the average request probability of the users currently associated with the F-AP \cite{Lu}. 

       Let $D _{i}^{m}$ denote the real number of requests of content $i$ in F-AP $m$, $D^{m}=\sum_{i=1}^{I}D_{i}^{m}$ the total number of requests in F-AP $m$, and $\psi _{i}^{m} \in \left \{ 0, 1 \right \}$ the cache status of content $i$ in F-AP $m$. $\psi _{i}^{m} = 1$ if content $i$ is stored in F-AP $m$, and $\psi _{i}^{m} = 0$ otherwise. The cache hit rate of F-AP $m$ is defined as the ratio of the number of cache hits to the total requests, and is utilized to evaluate the caching performance, which can be expressed as follows:
        \begin{equation}
        h^{m} = \frac{1}{D^{m}}\sum_{i=1}^{I} \psi _{i}^{m}\cdot D _{i}^{m}.
       \end{equation}
       With the content popularity in each F-AP determined, each F-AP can cache the most popular contents in its storage space.

       This paper aims to find a content popularity prediction policy with a high accuracy to maximize the cache hit rate in every F-AP considering both user mobility and regional variations of content popularity.

    \section{Proposed Clustered Federated Learning Based Popularity Prediction Policy}
    In this section, we propose a CFL based content popularity prediction policy, including local popularity prediction, mobile popularity prediction and popularity integration.  The proposed policy utilizes user features and content features, and can accurately predict and dynamically update the content popularity with low communication overhead.
    
    \subsection{Policy Description}
    
      1) \textit{Local popularity prediction}:
      The content request pattern reflects the user-content interaction,
      and the hidden representations of users and contents are responsible for the interaction \cite{FeatureInter}.
      In order to learn the hidden representations of local users and contents which are helpful to capture user-content interaction, DCNN model is introduced to produce deep latent features of local users and contents from initial features constructed by neighbor selection. Then, the request probability can be predicted by merging the deep latent features of local users and contents.
      To distinguish regional variations of local popularity, CFL is employed to specialize the model parameters of DCNN of each F-AP by grouping F-APs with similar regional types into the same cluster.
      Bandwidth resource can also be saved because only model parameters need to be transmitted in CFL.
      The local popularity in each F-AP is the integration of the predicted request probability and the activity levels of local users.

      2) \textit{Mobile popularity prediction}:
      Due to the random mobility of mobile users as well as privacy reasons, in order to predict the mobile popularity in a given F-AP, user preference learning is implemented independently by each mobile user in an offline manner.
      With all the currently associated mobile users sending the request probability predicted via user preference and content information, mobile popularity of contents in a given F-AP can be obtained.

      3) \textit{Popularity integration}:
      The content popularity in a given F-AP can be obtained by integrating the content popularities in terms of local users and mobile users.
      By monitoring the request records of local users as well as the user mobility of mobile users, the content popularity can be dynamically updated to maximize the cache hit rate.
      
%

      \subsection{Local Popularity Prediction}
      The procedure of local popularity prediction consists of the following three steps: neighbor selection based feature construction, DCNN model construction and CFL based model specialization.
      
      1) \textit{Neighbor selection based feature construction}: 
      To construct initial features of local users or contents, it is meaningful to take the information of users with similar preference or the information of contents requested by similar users into account \cite{Neighbor1}. For example, if a user prefers to movies which are favored mostly by teenagers, then the user's feature may be relevant to the information of teenagers. Therefore, we adopt a similarity computation method enhanced by inverse request frequency of users (IRFU) and inverse request frequency of contents (IRFC) to select neighbor set \cite{Neighbor2}. 
      
      Based on Euclidean distance, the similarity between two local users enhanced by IRFU can be computed by the rating of contents they both have requested according to \cite{Neighbor2}. Defined as the logarithmic transformation of the ratio of the total users to the users who have requested a certain content, IRFU reflects the discrimination of contents when computing the similarity between two local users.
      By utilizing IRFC of users, the similarity between two contents in a given F-AP can be computed analogously.
      Given similarities within the scope of a given F-AP $m$, the neighbor set $\mathcal{N}(l_{u}^{m})$ of  $l_{u}^{m}$ and $\mathcal{N}^{m}(i)$ of content $i$ are selected from the top $T$ most similar neighbors respectively.
      
      In order to incorporate neighbor information with self information, we propose to generate initial features via information integration.
      Let $\boldsymbol\xi_{u}^{m}\in [0,1]^{d_{U}}$ and $\boldsymbol\zeta_{i}\in [0,1]^{d_{I}}$ denote the information vectors of user $l_{u}^{m}$ and content $i$ respectively, where $d_{U}$ is the dimension of user information space and $d_{I}$ is the dimension of content information space. The initial feature vector $\boldsymbol x_{u}^{m}$ of user $l_{u}^{m}$ is computed as the weighted sum of the self information vector and the average information vector of neighbor set:
       \begin{equation}
       \boldsymbol x_{u}^{m}= w_{\mathcal{N}}\cdot \boldsymbol\xi_{u}^{m} + (1-w_{\mathcal{N}})\cdot\frac{1}{\left | \mathcal{N}(l_{u}^{m}) \right |} \sum_{l_{v}^{m} \subseteq \mathcal{N}(l_{u}^{m})}\boldsymbol\xi_{v}^{m},
       \end{equation}
      where $w_{\mathcal{N}}$ denotes the weight factor of self information in initial features. The initial feature vector $\boldsymbol{\chi}_{i}^{m}$ of content $i$ in F-AP $m$ is computed in a similar way:
       \begin{equation}
       \boldsymbol{\chi}_{i}^{m}= w_{\mathcal{N}}\cdot \boldsymbol\zeta_{i} + (1-w_{\mathcal{N}})\cdot\frac{1}{\left | \mathcal{N}^{m}(i) \right |} \sum_{j\subseteq \mathcal{N}^{m}(i)}\boldsymbol\zeta_{j}.
       \end{equation}

       
       
       2) \textit{DCNN model construction}: 
       Given initial features of local users and contents, a DCNN model is leveraged to learn the hidden representations by producing deep latent features in a given F-AP. 
       Motivated by \cite{DMF}, two feature transforming functions are constructed in DCNN to produce deep latent features of local users and contents respectively.
       Let $F_{U}:\mathbb{R}^{d_{U}} \mapsto \mathbb{R}^{H}$ and $F_{I}:\mathbb{R}^{d_{I}} \mapsto \mathbb{R}^{H}$ denote the feature transforming functions for local users and contents respectively. The deep latent features of local user $l_{u}^{m}$ and content $i$ in F-AP $m$ can be computed as $F_{U}(\boldsymbol x_{u}^{m}, \Theta_{U}^{m})$ and $F_{I}(\boldsymbol{\chi}_{i}^{m}, \Theta_{I}^{m})$, where $\Theta_{U}^{m}$ and $\Theta_{I}^{m}$ denote the parameters of the feature transforming functions in F-AP $m$.
        Let $\hat{p}_{u,i}^{m}$ denote the request probability of local user $l_{u}^{m}$ for content $i$. Then, it can be predicted by the inner product of the deep latent features as follows: 
       \begin{equation}
       \hat{p}_{u,i}^{m}=\mathrm{sigmoid}\left ( F_{U}(\boldsymbol x_{u}^{m}, \Theta_{U}^{m})^{T}\cdot F_{I}(\boldsymbol{\chi}_{i}^{m}, \Theta_{I}^{m}) \right ).
       \end{equation}
       
       We introduce the multilayer perceptron (MLP) as the feature transforming function in DCNN model, which is effective to learn the hidden representaions between input and output. Compared with the latent features generated from the observed request data \cite{pLSA1}, deep latent features generated from initial features contain the hidden representations, which contributes to capturing the user-content interaction.
       
       Let $\mathcal{A}^{m}=\left \{ (\boldsymbol x_{u}^{m}, \boldsymbol{\chi}_{i}^{m}, y_{u,i}^{m}), l_{u}^{m} \in \mathcal{L}^{m} , i \in \mathcal{I} \right \}$ denote the set of training samples of F-AP $m$, where $y_{u,i}^{m}$ is the binary request label. $y_{u,i}^{m}=1$ if local user $l^{m}_{u}$ has requested content $i$, otherwise $y_{u,i}^{m}=0$. The binary cross entropy (BCE) loss is utilized as the loss function for the optimization of parameters, which can be expressed as follows:
       \begin{equation}
       \eta = -[y_{u,i}^{m}\cdot \text{log}\hat{p}_{u,i}^{m} + (1-y_{u,i}^{m})\cdot \text{log}(1-\hat{p}_{u,i}^{m})].
       \end{equation}
       The DCNN model is trained by adjusting the parameters $\Theta_{U}^{m}$ and $\Theta_{I}^{m}$ through Adam optimizer with learning rate decaying.
       
       As the predicted request probability above is essentially the probability conditional on a sended request, it is necessary to take user activity level into account \cite{act1}. The activity level of local user $l^{m}_{u}$ (i.e., the probability that local user $l^{m}_{u}$ sends a request) is defined as $s^{m}_{u} = \left | R^{m}_{u} \right | /  \sum_{l_{v}^{m} \in \mathcal{L}^{m}} \left | R^{m}_{v} \right |$, where $\left | R^{m}_{u} \right |$ is the number of requests of local user $l_{u}^{m}$. According to the theory of total probability, local popularity of content $i$ in F-AP $m$ can be expressed as:
       \begin{equation}
       \hat{P}_{i}^{m}= \sum_{l_{u}^{m} \in \mathcal{L}^{m}} s^{m}_{u} \hat{p}_{u,i}^{m}.
       \end{equation}

       3) \textit{CFL based model specialization}:
       With training data stored in F-APs in a distributed manner, computational tasks for training a single global model can be distributed from the cloud server to F-APs via conventional FL \cite{FL}.
       Moreover, only model parameters need to be transmitted to the cloud server instead of all the training data, resulting in tremendous bandwidth resource savings.
       However, due to the diverse regional types of F-APs, local popularity in different F-APs varies. For example, users served by F-APs located near stadium are more likely to request contents related with sports, while users located at school tend to request educational contents. 
       In this situation, only one single global model obtained from conventional FL can hardly predict local popularity accurately, as all the F-APs with incongruent data distributions are treated equally.

       In order to solve the aforementioned problem with incongruent data distributions, CFL is adopted \cite{CFL}.
       CFL not only retains the advantage of saving bandwidth resource, but also adaptively splits up F-APs with incongruent data distributions into separated clusters without the prior knowledge of cluster number. Thus, F-APs with similar regional types (i.e., similar data distributions) can benefit from one another's training data in the same cluster, arriving at more specialized models with a higher accuracy.
       
       Let $\Phi =\left \{ \phi_{1}, \phi_{2},\dots,\phi_{g} ,\dots,\phi_{G} \right \}$ denote the final clustering result generated by CFL, where $\phi_{g}$ is a cluster containing at least one F-AP.
       The clustering result $\Phi$ can be obtained after $G-1$ valid bipartitions.  
       Let $\theta$ denote the parameter of DCNN model. 
       CFL will recursively split up F-APs and optimize model parameters in a top-down way. In the beginning, CFL is applied in the initial set of F-APs $\mathcal{M}$ with initial model parameter $\theta_{\mathcal{M}}$. Then, CFL will perform FL according to Algorithm \ref{alg:fl} with $\phi \equiv \mathcal{M}$,
       where $\varepsilon_{1}$ is the coefficient of FL stopping criteria.
       \begin{algorithm}[t]
    	\renewcommand{\algorithmicrequire}{\textbf{Input:}}
    	\caption{Procedure of FL in CFL}
    	\label{alg:fl} 
    	\begin{algorithmic}
    		\Require {initial parameter $\theta_{\phi}$, set of F-APs $\phi$, $\varepsilon _{1}>0$;}
    		\Repeat 
    			\For{F-AP $m \in \phi$ \textbf{in parallel}}
    				\State \underline{F-AP $m$ does:} 
    				\State $\theta _{m}\leftarrow \theta_{\phi}$;
    				\State $\Delta\theta_{m}\leftarrow \text{SGD}(\theta _{m},\mathcal{A}^{m})-\theta _{m}$;
    			\EndFor
    			\State \underline{The cloud server does:}
    			\State $\theta_{\phi} \leftarrow \theta_{\phi} + \sum _{m \in \phi}\frac{\left | \mathcal{A}^{m} \right |}{\left | \mathcal{A}^{\phi} \right |}\Delta\theta_{m}$;
    			
    		\Until{$\left \| \sum _{m \in \phi}\frac{\left | \mathcal{A}^{m} \right |}{\left | \mathcal{A}^{\phi} \right |} \Delta \theta_{m}  \right \|<\varepsilon_{1}$};
    		\State \Return $\theta_{\phi}$;
    	\end{algorithmic}
     \end{algorithm}
       After FL has converged, the FL solution $\theta^{*}_{\mathcal{M}}$ of $\mathcal{M}$ can be obtained. Then, the following clustering criteria is evaluated:
       \begin{equation}
       0\leq \underset{m \in \phi}{\max}\left \| \Delta \theta _{m} \right \|< \varepsilon _{2},
       \label{eq:criteria2}
       \end{equation}
        where $\varepsilon _{2}$ is the coefficient of the clustering criteria. If the clustering criteria is satisfied in $\mathcal{M}$, all F-APs are thought to be sufficiently close to their optimal model parameters. CFL terminates and $\theta^{*}_{\mathcal{M}}$ will serve as the solution for every F-AP in $\mathcal{M}$. If not satisfied, F-APs in $\mathcal{M}$ are thought to be incongruent and the pairwise cosine similarity $\alpha$ will be computed in $\mathcal{M}$ by utilizing F-APs' model weight updates:
       \begin{equation}
       \alpha _{m,n}=\alpha \left ( \Delta \theta _{m},\Delta\theta _{n} \right )=\frac{\left \langle  \Delta\theta _{m},\Delta\theta _{n}\right \rangle}{\left \| \Delta\theta _{m} \right \| \cdot  \left \| \Delta\theta _{n} \right \|}, \forall m,n \in \phi.
       \label{eq:sim}
       \end{equation}
       Given the pairwise similarity, the cloud server will split up the F-APs in $\mathcal{M}$ into two sub-clusters by minimizing the maximum similarity between F-APs from different sub-clusters:
       \begin{equation}
       \phi_{1},\phi_{2}\leftarrow \arg\underset{\phi_{1}\cup \phi_{2}=\phi}{\min}\left ( \underset{m \in \phi_{1}, n \in \phi_{2}}{\max} \alpha_{m,n} \right ).
       \label{eq:split}
       \end{equation}
       Then, F-APs will aggregate their model parameters within the newly seperated clusters by averaging their model weight updates. Let $\phi(m)$ denote the cluster containing F-AP $m$. Then, the aggregation operation can be expressed as:
       \begin{equation}
       \theta _{m}\leftarrow \theta_{\phi(m)} = \theta _{m}+ \frac{1}{\left | \phi(m) \right |} \sum_{n \in \phi(m)}\Delta \theta_{n}, \forall m \in \phi(m).
       \label{eq:update}
       \end{equation}
       
       If any F-AP has not converged to its optimal model parameter, CFL will be recursively reapplied in each of the two sub-clusters based on their own updated model parameters $\theta_{\phi_{1}}$ and $\theta_{\phi_{2}}$ seperately. When none of the sub-clusters violates the clustering criteria in (\ref{eq:criteria2}), all clusters of the congruent F-APs $\Phi$ have been identified, and each F-AP will obtain a more specialized DCNN model with a higher accuracy. The detailed procedure of CFL based model specialization is presented in Algorithm \ref{alg:cfl}.
       
      \begin{algorithm}[t]
    	\renewcommand{\algorithmicrequire}{\textbf{Input:}}
    	\renewcommand{\algorithmicensure}{\textbf{Output:}}
    	\caption{CFL based model specialization}
    	\label{alg:cfl}
    	\begin{algorithmic}[1]
    		\Require {initial model parameter $\theta_{\mathcal{M}}$ and $\varepsilon _{1},\varepsilon _{2}> 0$;} 
    		\Ensure {specialized model parameters $\theta_{m}$ for each F-AP $m$;}
    		\State \textbf{init:} set initial cluster $ \Phi =\left \{ \mathcal{M} \right \}$, set initial model $\theta _{m}\leftarrow \theta _{\mathcal{M}},\forall m \in \mathcal{M}$, set initial weight update $\Delta \theta _{m}\leftarrow 0, \forall m \in \mathcal{M}$;
    		\While{not converged} 
    			\For{F-AP $m=1, 2, \dots, M$ \textbf{in parallel}}
    				\State \underline{F-AP $m$ does:} 
    				\State $\Delta\theta_{m}\leftarrow \text{SGD}(\theta _{m},\mathcal{A}^{m})-\theta _{m}$;
    			\EndFor
    			\State \underline{The cloud server does:}
    			\State $\Phi_{temp}\leftarrow \Phi$;
    			\For{$\phi \in \Phi$}
    				\State $\Delta \theta_{\phi} \leftarrow \sum _{m \in \phi}\frac{\left | \mathcal{A}^{m} \right |}{\left | \mathcal{A}^{\phi} \right |}\Delta\theta_{m}$;
    				\If{$\left \| \Delta \theta_{\phi}  \right \|<\varepsilon_{1}$ \textbf{and} $ \underset{m \in \phi}{\max}\left \| \Delta \theta _{m} \right \| > \varepsilon _{2}$}
    				\State Compute similarity $\alpha$ according to (\ref{eq:sim});
    				\State Split up $\phi$ into $\phi_{1}$ and $\phi_{2}$ according to (\ref{eq:split});
    				\State $\Phi_{temp}\leftarrow (\Phi_{temp} \setminus \phi)\cup \phi_{1}\cup \phi_{2}$;
    				\EndIf
    			\EndFor
    			\State $\Phi \leftarrow \Phi_{temp}$;
    			\For{F-AP $m=1, 2, \dots, M$ \textbf{in parallel}}
    				\State \underline{F-AP $m$ does:} 
    				\State Update $\theta _{m}$ according to (\ref{eq:update});
				\EndFor
    		\EndWhile
    	\end{algorithmic}
     \end{algorithm}

      \subsection{Mobile Popularity Prediction}
       Corresponding to the content information vector $\boldsymbol\zeta_{i}\in [0,1]^{d_{I}}$, let $\boldsymbol{a}_{u}=[a_{u1},a_{u2},\dots,a_{ud_{I}}]^{T}$ denote the user preference vector of mobile user $k_{u}$, which represents the preference for different content information. The learning process of user preference will be carried out independently by mobile users.
       
       For mobile user $k_{u}$, the set of training samples $\mathcal{B}_{u}=\left \{ (\boldsymbol\zeta_{i},y_{u,i}), i \in \mathcal{I} \right \}$ is extracted from its local request records, where $y_{u,i}$ is the binary request label.
       $y_{u,i}=1$ if mobile user $k_{u}$ has requested content $i$, otherwise $y_{u,i}=0$. Let $\hat{q}_{u,i}$ denote the probability that mobile user $k_{u}$ has requested content $i$, which can be predicted based on content information and user preference as follows:
       \begin{equation}
		\hat{q}_{u,i}=p_{\boldsymbol{a}_{u}}(y_{u,i}=1\mid \boldsymbol\zeta_{i})=\mathrm{sigmoid}(\boldsymbol{a}_{u}\ast \boldsymbol\zeta_{i}).
      \end{equation}
      
      Given the predicted $\hat{q}_{u,i}$, the negative log-likelihood of $y_{u,i}$ is formulated as follows:
       \begin{equation}
		\ell(\boldsymbol{a}_{u})=-\frac{1}{\left | \mathcal{B}_{u} \right |}\sum _{(\boldsymbol\zeta_{i},y_{u,i}) \in \mathcal{B}_{u}}[y_{u,i}\text{log}\hat{q}_{u,i} + (1-y_{u,i})\text{log}(1-\hat{q}_{u,i})].
      \end{equation}
      The ``Follow The (Proximally) Regularized Leader''(FTRL-Proximal) algorithm is adopted to obtain the user preference of mobile users by minimizing $\ell(\boldsymbol{a}_{u})$ in an offline manner \cite{Ma}, and the user preference will be periodically updated when $\ell(\boldsymbol{a}_{u})$ exceeds a certain threshold.
      
      Due to privacy reasons, F-APs cannot access historical request records of mobile users. Without loss of generality, we assume the activity levels of mobile users in an F-AP are the same. Let $\mathcal{K}^{m}\subseteq \mathcal{K}$ denote the set of mobile users currently associated with F-AP $m$.
      All the currently associated mobile users only need to send the predicted request probability to the F-AP, which also contributes to the saving of bandwidth resource.
      Then, the predicted mobile popularity of content $i$ in F-AP $m$ can be computed as the average request probability of the currently associated mobile users:
       \begin{equation}
		\hat{Q}_{i}^{m}= \frac{1}{\left | \mathcal{K}^{m} \right |} \sum_{k_{u} \in \mathcal{K}^{m}} \hat{q}_{u,i}.
      \end{equation}

      \subsection{Popularity integration}
      In order to eliminate the differences between local popularity and mobile popularity, local popularity and mobile popularity of all the contents will be normalized respectively in each F-AP as follows: $\hat{P}_{i}^{m} \leftarrow \hat{P}_{i}^{m} / \sum _{j \in \mathcal{I}} \hat{P}_{j}^{m}$ and $\hat{Q}_{i}^{m} \leftarrow \hat{Q}_{i}^{m} / \sum _{j \in \mathcal{I}} \hat{Q}_{j}^{m}$.
      Then, the predicted popularity $\hat{pop}_{i}^{m}$ of content $i$ in F-AP $m$ can be calculated to be:
        \begin{equation}
        \hat{pop}_{i}^{m} = (1-w_{m})\cdot \hat{P}_{i}^{m} + w_{m} \cdot \hat{Q}_{i}^{m},
        \label{eq:pop}
       \end{equation}      
      where $w_{m}$ is defined as the ratio of the currently associated mobile users to the total users in F-AP $m$ for the purpose of tackling the varying number of mobile users caused by user mobility.
      By the integration of local and mobile popularities, both user mobility and regional variations of content popularity are properly considered in our proposed content popularity prediction policy.
      
      According to the above descriptions, the proposed popularity prediction policy not only dynamically updates the predicted content popularity by monitoring the request records of local users as well as the user mobility of mobile users, but also significantly reduces communication overhead.

    \section{Simulation Results}  
     To evaluate the performance of the proposed popularity prediction policy, simulations are performed based on data extracted from the MovieLens 1M Dataset \cite{Movie}. The dataset contains 1,000,209 ratings for approximately 3,900 movies created by 6,040 users. For each item, a user ID, a movie ID, a rating from 1 to 5 and a timestamp are included. Each record is regarded as a request of user for content. Demographic information of users is provided in the dataset, which includes gender, age, occupation and Zip-code. By utilizing one-hot encoding, the information of gender, age and occupation is transformed to a binary vector, which can serve as the user information. Furthermore, the provided genres of movies can be used as content information. In our simulations, we set the number of F-APs $M$ to 10, the size of neighbor set $T$ to 20 and the weight factor of self information $w_{\mathcal{N}}$ to 0.5, respectively. 
     Without considering mobile popularity, policies based on DCNN model obtained by CFL (DCNN-CFL), FL (DCNN-FL) and local learning (DCNN-LC), as well as the pLSA based policy \cite{pLSA1}, LFU and LRU are chosen as the benchmark policies.
     
        \begin{figure}[!t]
    	 \centering
    	 \includegraphics[width=8cm]{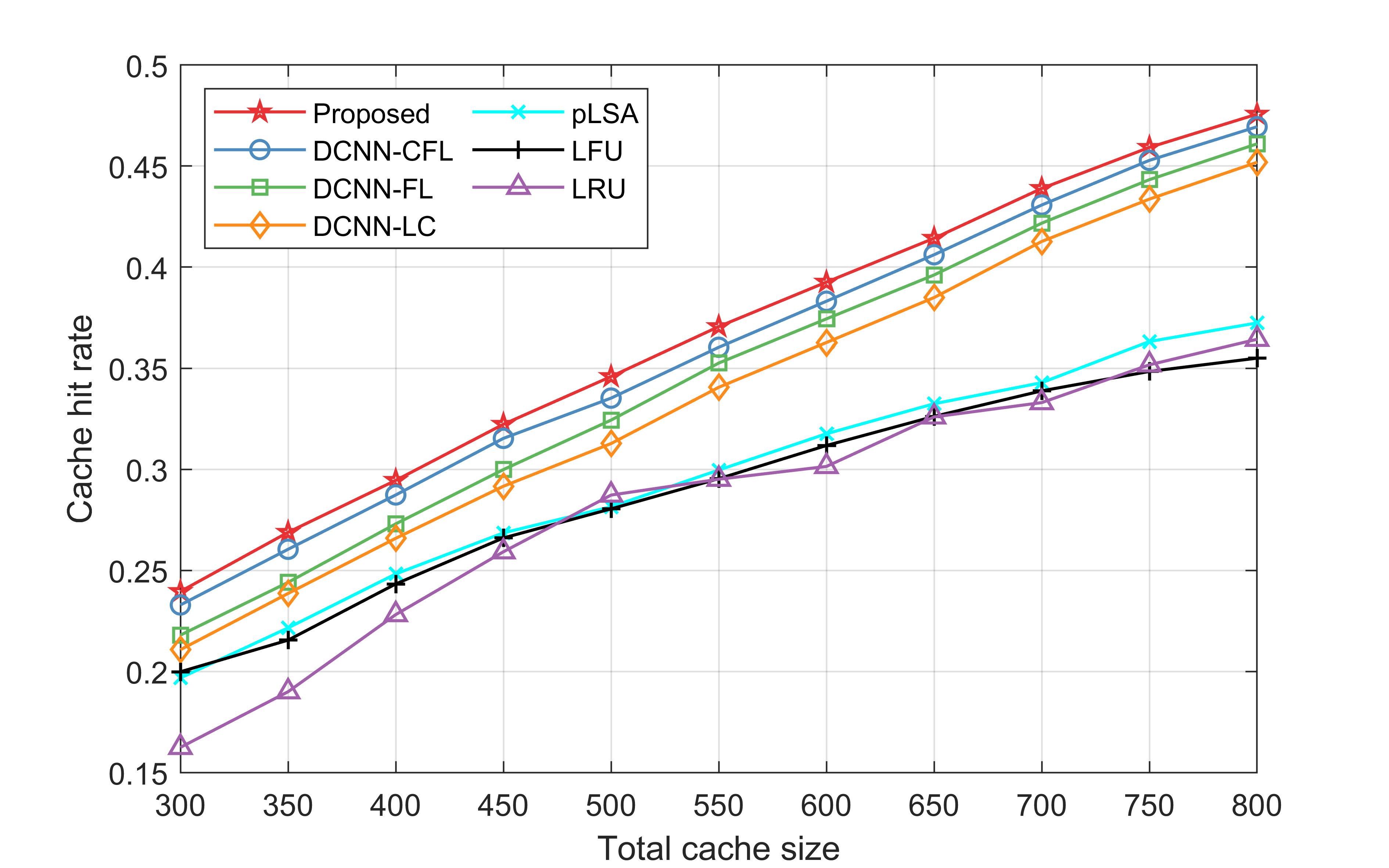} 
    	 \caption{Cache hit rate versus total cache size.}  \label{fig:size}
        \end{figure}
        
        \begin{figure}[!t]
    	 \centering
    	 \includegraphics[width=8cm]{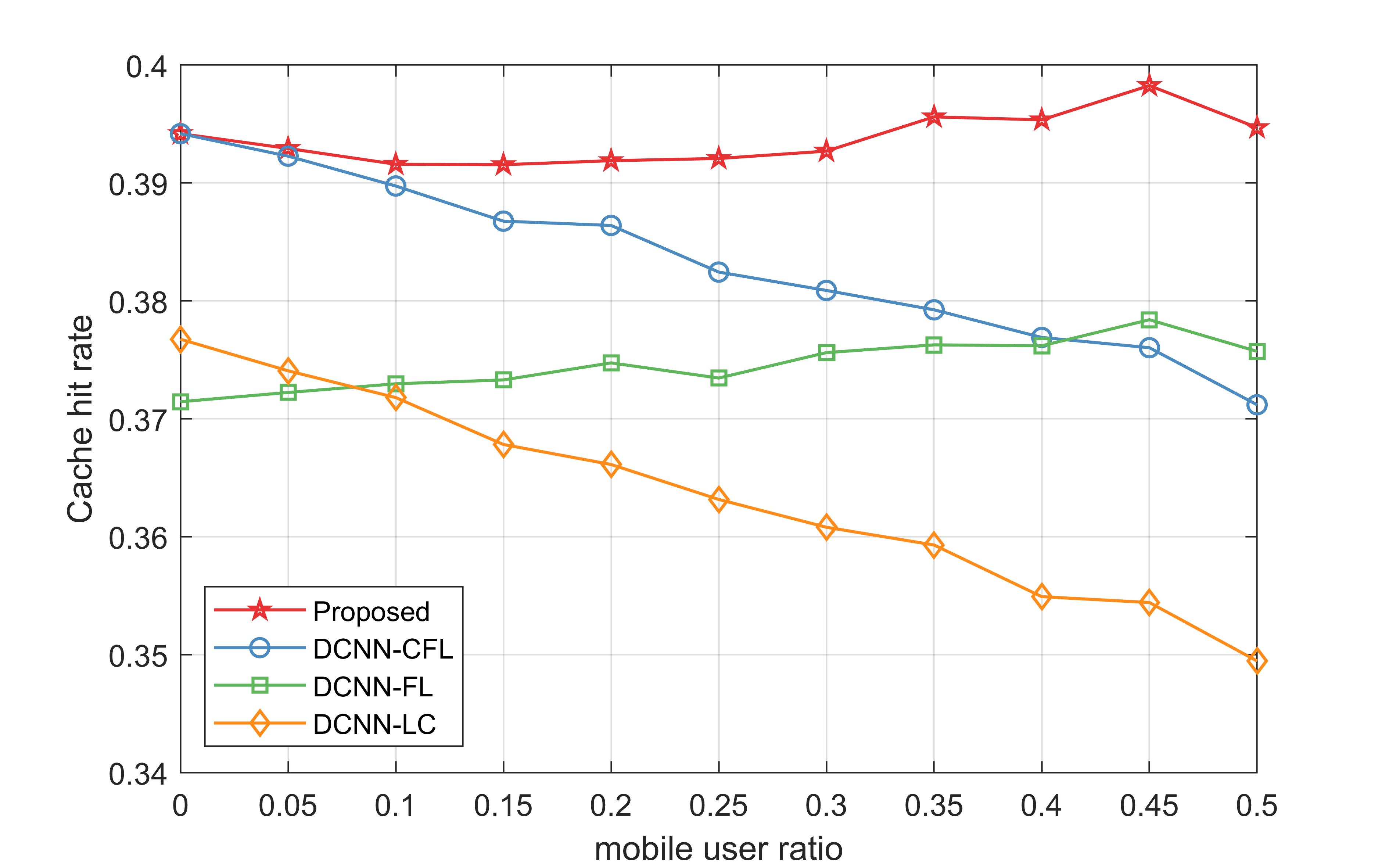} 
    	 \caption{Cache hit rate versus mobile user ratio.}  \label{fig:mur}
        \end{figure}
     In Fig. \ref{fig:size}, we show the cache hit rates of our proposed policy and the benchmark policies versus the total cache size with 25\% mobile users in each F-AP.
     It can be observed that the cache hit rates of our proposed policy and the DCNN model based policies are larger than those of the other benchmark policies.
     The reason is that through DCNN model, the hidden representations are learned by producing deep latent features from initial features, which are helpful to capture user-content interaction. Furthermore, neighbor information is taken into consideration to generate initial features.
     Whereas the pLSA based policy extracts latent features merely through historical request records via a probabilistic method, and LFU and LRU can hardly achieve satisfactory performance due to the neglect of content popularity.
     
     In Fig. \ref{fig:mur}, we show the cache hit rates of our proposed policy and the DCNN model based policies versus the mobile user ratio for a total cache size of 600. 
     It can be observed that without any mobile users in F-APs, the proposed policy (i.e., DCNN-CFL) outperforms the other two policies. The reason is that our proposed policy enables F-APs with similar regional types (i.e., with similar data distributions) to benefit from one another while predicting local popularity. DCNN-FL has the worst performance when no mobile users in F-APs, as all the F-APs with incongruent data distributions are treated equally by training one single global model.
     It can also be observed that as the mobile user ratio increases, our proposed policy and DCNN-FL maintain their original preformance level with little fluctuation, whereas the other two policies suffer from a performance degradation. The reason is that our proposed policy predicts content popularity by integrating mobile popularity which is updated dynamically according to the mobility and the preference of mobile users. The content popularity predicted by DCNN-FL is more general and is adaptive to the request of mobile users, leveraging a global model obtained by FL.
     Moreover, the cache hit rate of our proposed policy is larger than that of DCNN-FL for all mobile user ratios. The reason is that our proposed policy predicts content popularity by integrating both local and mobile popularities. 
     
     

    \section{Conclusions}
     In this paper, we have proposed a novel mobility-aware popularity prediction policy based on CFL in F-RANs.
     Our proposed policy can accurately predict and dynamically update content popularity
     even when users move randomly among F-APs. 
     The reason is that mobile popularity is considered by utilizing user preference learning.
     Specifically, we have proposed to utilize CFL to enable F-APs with similar regional types to benefit from one another, which provides a more specialized model with a higher accuracy for each F-AP and significantly reduces communication overhead. 
     Simulation results have shown that our proposed policy outperforms benchmark policies in terms of cache hit rate.

    \section*{Acknowledgments}
     This work was supported in part by the National Key Research and Development Program under Grant 2021YFB2900300, the National Natural Science Foundation of China under grant 61971129, and the Shenzhen Science and Technology Program under Grant KQTD20190929172545139.
    
  \balance
  \bibliographystyle{IEEEtran}
  \bibliography{manuscript-refer}  
    
\end{document}